\documentclass[10pt,twocolumn,letterpaper]{article}

\usepackage{cvpr}              

\usepackage[dvipsnames]{xcolor}

\usepackage{graphicx}
\usepackage{booktabs}
\usepackage{epsfig}
\usepackage{enumitem}
\usepackage{amsmath}
\usepackage{amssymb}
\usepackage{xcolor} 
\usepackage{bm}
\usepackage{overpic}
\usepackage{multirow}
\usepackage{caption}
\usepackage{setspace}
\usepackage{adjustbox} 

\usepackage{times}
\usepackage{epsfig}
\usepackage{graphicx}
\usepackage{amsmath}
\usepackage{amssymb}
\usepackage{bm}
\usepackage{overpic}
\usepackage{multirow}
\usepackage{booktabs}
\usepackage{enumitem}
\usepackage{soul}
\usepackage{caption}
\usepackage{cite}

\definecolor{cvprblue}{rgb}{0.21,0.49,0.74}
\usepackage[pagebackref,breaklinks,colorlinks,citecolor=cvprblue]{hyperref}

\def\paperID{213} 
\def\confName{3DV\xspace}
\def\confYear{2025\xspace}

\title{Fully-Geometric Cross-Attention for Point Cloud Registration}

\author{Weijie Wang\\
University of Trento\\
{\tt\small weijie.wang@unitn.it}
\and
Guofeng Mei\thanks{ denotes the corresponding author.}\\
Fondazione Bruno Kessler\\
{\tt\small gmei@fbk.eu}
\and
Jian Zhang\\
University of Technology Sydney\\
{\tt\small Jian.Zhang@uts.edu.au}
\and
Nicu Sebe\\
University of Trento\\
{\tt\small niculae.sebe@unitn.it}
\and
Bruno Lepri\\
Fondazione Bruno Kessler\\
{\tt\small lepri@fbk.eu}
\and
Fabio Poiesi\\
Fondazione Bruno Kessler\\
{\tt\small poiesi@fbk.eu} \\
}

\newcommand{\ourmethod}{FLAT\xspace}
\begin{document}
\maketitle
Point cloud registration approaches often fail when the overlap between point clouds is low due to noisy point correspondences. 
This work introduces a novel cross-attention mechanism tailored for Transformer-based architectures that tackles this problem, by fusing information from coordinates and features at the super-point level between point clouds.
This formulation has remained unexplored primarily because it must guarantee rotation and translation invariance since point clouds reside in different and independent reference frames.
We integrate the Gromov–Wasserstein distance into the cross-attention formulation to jointly compute distances between points across different point clouds and account for their geometric structure.
By doing so, points from two distinct point clouds can attend to each other under arbitrary rigid transformations.
At the point level, we also devise a self-attention mechanism that aggregates the local geometric structure information into point features for fine matching. 
Our formulation boosts the number of inlier correspondences, thereby yielding more precise registration results compared to state-of-the-art approaches. 
We have conducted an extensive evaluation on 3DMatch, 3DLoMatch, KITTI, and 3DCSR datasets. 
Project page: \url{https://github.com/twowwj/FLAT}.
    
\section{Introduction}
\label{sec:intro}
\thispagestyle{empty}
Point cloud registration aims to determine a relatively rigid transformation that aligns two partially observed point clouds. 
This is important in various applications, including 3D printing \cite{xie2023bayesian, nie2019hgan, chang20243d, nie2024t2td}, robotics \cite{kim2018slam, wang2024uvmap, li2025freeinsert}, and autonomous driving \cite{brightman2023point}.
In recent registration advances, deep learning-based methods have outperformed hand-crafted ones in both efficiency and accuracy \cite{choy2019fully, GEDI}.

Learning-based point cloud registration 
can be categorized into \textit{correspondence-free}~\cite{aoki2019pointnetlk, huang2020feature, xu2021omnet, mei2022augfree} or \textit{correspondence-based}~\cite{GEDI, choy2020deep, bai2021pointdsc, huang2021predator, yew2022regtr, corsetti2023revisit, wang2024zeroregzeroshotpointcloud}.
The former aims to minimize the difference between the global features of two point clouds.
However, global features may hinder these approaches in handling partially overlapping scenes~\cite{zhang2020deep, choy2020deep}. 
The latter aims to detect keypoints, compute local descriptors, find correspondences, and estimate the rigid transformation~\cite{huang2021predator, mei2023unsupervised}.
Correspondences can be defined at the point level or at the distribution level~\cite{GEDI,wang2023roreg}.
Point-level correspondences may be noisy in the case of point clouds with different densities of points and/or with geometrically repetitive and uninformative local patterns (e.g.~flat surfaces)~\cite{xu2022glorn,mei2023overlap}.
Distribution-level correspondences are designed to align point clouds with varying densities without establishing point-level correspondences, however, they fall short in handling point clouds with low overlaps~\cite{mei2022overlap,huang2022unsupervised}.
There also exist methods that replace keypoint detection by downsampling input point clouds into super-points in order to make computation efficient~\cite{yu2021cofinet}.
Super-points are matched between point clouds to find correspondences, which are in turn propagated at point-level to build dense point correspondences.
Furthermore, self-attention and cross-attention in Transformers~\cite{vaswani2017attention} can be used to incorporate global information (context) into features~\cite{mei2023unsupervised, huang2021predator, yu2021cofinet, qin2022geometric}, thus producing distinctive features to register point clouds more accurately.
Self-attention is performed in the coordinate space to encode the transformation-invariant geometric structure from each point cloud \cite{qin2022geometric}.
Cross-attention is performed in the feature space to model the geometric consistency across point clouds, by allowing information from one point cloud to be attended by another \cite{qin2022geometric}.

We argue that previously proposed cross-attention formulations only employ point feature information, overlooking coordinate information.
Although cross-attention in the feature space is effective in improving correspondence quality between super-points, point-level correspondences remain rather noisy (we will show this experimentally).
Intuitively, if we only consider feature information, disregarding their location, there could be situations where similar objects in different locations have similar geometric structures, hence similar features. 
This can produce incorrect correspondences. 
By enriching the cross-attention with coordinate information, we can encourage the network to explicitly learn corresponding geometric structures across point clouds, thus promoting feature distinctiveness.

To this end, in this paper we present a \textbf{f}u\textbf{l}ly-geometric cross-\textbf{at}tention formulation, \ourmethod for short, followed by overlap-constrained clustering to learn accurate correspondences for point cloud registration.
Unlike GeoTransformer~\cite{qin2022geometric}, which only considers geometric relationships within a single point cloud. Our method, \ourmethod, uses geometric cross-attention to incorporate both source and target relationships. This provides comprehensive scene information, overcoming the limitations of partial scene representation in overlapped point clouds. In addition, we develop cross-spatial invariant geometric features and use the Gromov-Wasserstein distance to measure discrepancies across different metric spaces, such as pose differences.
Our cross-attention is significantly more challenging to achieve, as measures of distance and angle used in GeoTransformer are no longer applicable because they are not invariant to rigid transformations with respect to different reference frames.
To fix this gap, we introduce two new metrics, one for measuring the pair-wise distance and the other for determining the triplet-wise angle that can be computed between two different point clouds.
As these two metrics are invariant to rigid transformation, our geometry-enhanced attention can efficiently exchange geometric structural information between point clouds, leading to more reliable correspondences, even in scenarios with low overlaps.
Our registration approach follows a coarse-to-fine correspondence prediction strategy, identifying approximate matches using super-points and refining them by expanding them to patches (i.e.~sets of points defined in the neighborhood of a super-point).
Moreover, unlike previous techniques, we employ a distance-weighted self-attention to inject the local location information to further improve the distinctive of the local features.
We evaluate our method on four popular benchmarks:
3DMatch \cite{zeng20173dmatch} and 3DLoMatch \cite{huang2021predator} (indoors), KITTI \cite{geiger2012we} (outdoors), and cross-source 3DCSR~\cite{huang2021comprehensive} (indoors).
The results show that \ourmethod outperforms previous approaches~\cite{huang2021predator,yu2021cofinet,qin2022geometric}.
In summary, our contributions are:
\begin{itemize}[noitemsep,topsep=0pt]

\item We introduce a geometry-enhanced cross-attention mechanism alongside a distance-weighted self-attention approach, aiming to refine the learning of accurate correspondences for point cloud registration.
\item Our proposed geometry-enhanced cross-attention effectively integrates a transformation invariant geometric structure, enabling the model to learn distinctive features and emphasize the overlapping regions between point clouds.
\item Leveraging local self-attention, which is grounded in coordinate relative distances, we generate distinctive features that enhance fine matching capabilities.
\end{itemize}

\section{Related Work}

\noindent\textbf{Correspondence-based Registration.}
Correspondence-based methods typically involve two main steps: feature extraction and correspondence estimation through feature matching. 
They perform outlier rejection and robust estimation of the rigid transformation. 
FCGF~\cite{choy2019fully}, RGM~\cite{fu2021robust}, and GeDi~\cite{GEDI} are examples of methods used to extract discriminative features. 
For correspondence prediction, RPMNet~\cite{yew2020rpm} performs feature matching by integrating the Sinkhorn algorithm into a network that generates soft feature correspondences.
IDAM~\cite{li2019iterative} incorporates both geometric and distance features in the iterative matching process. 
To reject outliers, DGR~\cite{choy2020deep} and 3DRegNet~\cite{pais20203dregnet} use networks for inlier prediction.
Correspondence-based techniques can be further classified into two groups based on the strategy they employ to extract correspondences~\cite{qin2022geometric}. 
The methods in the first group aim to identify repeatable keypoints \cite{bai2020d3feat, huang2021predator} and develop discriminative descriptors for those keypoints \cite{ao2021spinnet, wang2022you, wang2023roreg}. 
The effectiveness of keypoint detection methods may be limited in scenarios with uneven point density or similar local structures. 
Repetitive local structures are typically present in indoor settings, where featureless flat surfaces can occupy a significant portion of the visual field. 
The methods in the second group aim to retrieve correspondences without detecting keypoints by examining all possible matches~\cite{yu2021cofinet, qin2022geometric}. 
They first downsample the point clouds into super-points and then match them by examining whether their neighborhoods (patches) overlap~\cite{mei2021point, zhang2022patchformer, yu2021cofinet}. 
The accuracy of dense point correspondences relies on the accuracy of super-point matches~\cite{qin2022geometric}. 
When dealing with low-overlapping point clouds, the super-point matching mechanism merely exchanges information from the feature spaces, which contain only partial structural information. 
This limitation emerges when compared to the original 3D scenes contained within the point cloud pair, leading to false matches. 
The points of a patch tend to have similar features, which can challenge dense correspondence prediction. 
To overcome these limitations, \ourmethod merges the advantages of both cross-geometric structures and feature relationships, enhancing the accuracy of super-point matching. 
We introduce a distance-enhanced local self-attention mechanism that generates point-level features, improving dense matching capabilities.

\noindent\textbf{Transformer on Registration.}
Transformer attention~\cite{vaswani2017attention} has recently been successful in point cloud tasks due to its ability to learn long-range dependencies and invariance to input token permutations.
Using the Transformer has been shown to enhance point cloud registration performance effectively.
DCP~\cite{wang2019deep} applies standard cross-attention to highlight similarities of matched points across two point clouds for soft correspondence generation.
The Geometric Transformer~\cite{qin2022geometric} aims to improve feature matching accuracy by enhancing the effectiveness of self-attention in acquiring geometric information.
Their method encodes both pairwise distances and triplet-wise angles, allowing it to handle low-overlap scenarios while remaining invariant to rigid transformations.
REGTR~\cite{yew2022regtr} integrates the Transformer~\cite{vaswani2017attention} into a network that generates soft correspondences from local features, allowing feature matching for point clouds with partial overlaps.
Predator~\cite{huang2021predator} and PRNet~\cite{wang2019prnet} apply Transformer to detect points in the overlap region and use the features of the detected points to generate matches.
Several of these methods fail to consider the fact that multiple regions within a point cloud may display similar structures, which can limit the effectiveness of standard cross-attention when dealing with comparable local structures.
Motivated by this, we propose an improved cross-attention that incorporates transformation-invariant geometric structure into learned features to better highlight overlap regions of both point clouds.
\section{Our approach}

\begin{figure*}[t]
\centering  
\includegraphics[width=.99\textwidth]{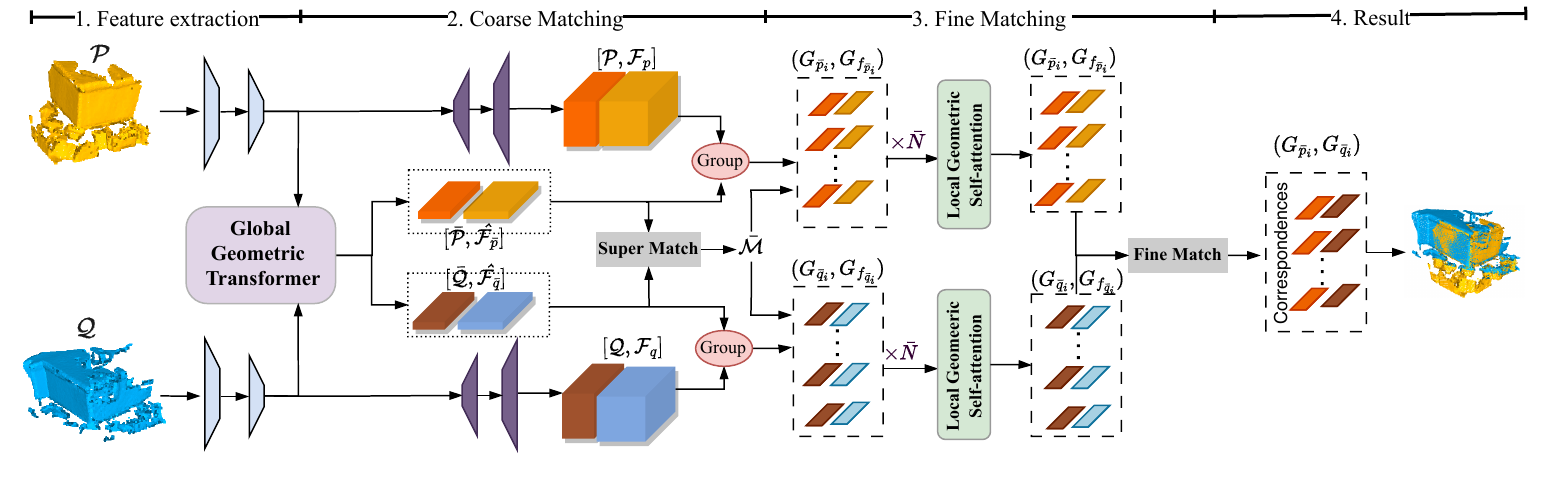}
\vspace{-3mm}
\caption{
The encoder downsizes the input point clouds and generates super-points with associated features. 
The global geometric Transformer injects geometric information into learned features. 
Coarse matching of the super-points is carried out between the two downsampled inputs. 
The local geometric Transformer is utilized to generate distinctive local features, which enables the prediction of fine-level correspondences between the inputs. 
Finally, the rigid transformation is estimated from the fine-level correspondences.}
\label{fig:frame}
\end{figure*}

\subsection{Overview}\label{sec:probf}
Point cloud registration refers to recover a transformation $\bm{T}\in SE(3)$ that aligns the source set $\bm{\mathcal{P}}=\{\bm{p}_i \in\mathbb{R}^{3}|i = 1, 2, ..., N\}$ to the target set $\bm{\mathcal{Q}}=\{\bm{q}_j \in\mathbb{R}^{3}|j = 1, 2, ..., M\}$.
$N$ and $M$ are the number of points in $\bm{\mathcal{P}}$ and $\bm{\mathcal{Q}}$, respectively. $\bm{T}$ can be calculated using correspondences between $\bm{\mathcal{P}}$ and $\bm{\mathcal{Q}}$. 
Fig.~\ref{fig:frame} shows \ourmethod's pipeline, which adopts a hierarchical paradigm to predict correspondences in a coarse-to-fine manner. 
Given the pair of point clouds $\bm{\mathcal{P}}$ and $\bm{\mathcal{Q}}$, the encoder aggregates the raw points into superpoints $\bar{\bm{\mathcal{P}}}$ and $\bar{\bm{\mathcal{Q}}}$, while jointly learning their associated characteristics $\bm{\mathcal{F}}_{\bar{p}}$ and $\bm{\mathcal{F}}_{\bar{q}}$. 
The full geometric Transformer block updates the features as $\hat{\bm{\mathcal{F}}}_{\bar{p}}$ and $\hat{\bm{\mathcal{F}}}_{\bar{q}}$. 
A \textit{Coarse Matching} module is applied to extract super-point correspondences whose neighboring local patches overlap with each other. 
The decoder transforms the features into per-point features $\bm{\mathcal{F}}_{p}$, $\bm{\mathcal{F}}_{q}$.
Then, we apply a local geometric self-attention on each patch to refine the features.
Lastly, a \textit{Fine Matching} module extracts cluster-level correspondences, which are then used to estimate the transformation.   

\subsection{Feature Extraction}
\noindent\textbf{Encoder.}
We use KPConv-FPN \cite{thomas2019kpconv} to downsample $\bm{\mathcal{P}}$ and $\bm{\mathcal{Q}}$ into super-points $\bar{\bm{\mathcal{P}}}=\{\bar{\bm{p}}_i \in\mathbb{R}^{3}|i = 1, 2, ..., \bar{N}\}$ and $\bar{\bm{\mathcal{Q}}}=\{\bar{\bm{q}}_j \in\mathbb{R}^{3}|j = 1, 2, ..., \bar{M}\}$, and to extract associated point-wise features $\bm{\mathcal{F}}_{\bar{p}}=\{\bm{f}_{\bar{p}} \in\mathbb{R}^{b}|i = 1, 2, ..., \bar{N}\}$ and $\bm{\mathcal{F}}_{\bar{q}}=\{\bm{f}_{\bar{q}_j} \in\mathbb{R}^{b}|j = 1, 2, ..., \bar{M}\}$, respectively, with $b=256$.
KPConv-FPN consists of a series of ResNet-like blocks and stridden convolutions.

\vspace{-2mm}

\subsection{Full Geometric Transformer} 

\noindent\textbf{Global Geometric Self-attention.}
We use the Geometric self-attention as it is implemented in GeoTransformer~\cite{qin2022geometric}.



\noindent\textbf{Global Geometric Cross-attention.}
We develop a new fully-geometric cross-attention strategy that can identify global correlations between super-points in two point clouds, based on both feature and coordinate information.
Given two super-points $\bar{\bm{p}}_i\in \bar{\bm{\mathcal{P}}}$ and $\bar{\bm{q}}_j\in \bar{\bm{\mathcal{Q}}}$, the cross-attention output $\bm{\hat{f}}_{\bar{p}_i}\in\hat{\bm{\mathcal{F}}}_{\bar{p}}$ for $\bm{\bar{p}}_i$ can be obtained by computing the weighted sum of all projected input features:
\begin{equation}
    \bm{\hat{f}}_{\bar{p}_i} = \sum_{j=1}^{\bar{M}}\alpha_{ij}\bm{f}_{\bar{q}_j}\bm{W}^V,
\end{equation}
where the weight coefficient $\alpha_{ij}$ is calculated using row-wise softmax on the attention scores as:
\begin{equation}
    \alpha_{ij} = \mbox{softmax}\left(  {\bm{f}_{\bar{p}_i}\bm{W}^Q\left(\bm{f}_{\bar{q}_j}\bm{W}^K+\bm{r}_{ij}\bm{W}^R\right)^\top}\big/\sqrt{b}\right),
\end{equation}
where $\bm{W}^Q,\bm{W}^K,\bm{W}^R,\bm{W}^V$ correspond to the projection of queries, keys, values, and geometric structure embeddings, respectively.
$\bm{r}_{ij}$ is the embedding of the cross-geometric structure of distance and angle that we compute as follows. 

\begin{figure}[t]
\centering  
\includegraphics[width=0.95\columnwidth]{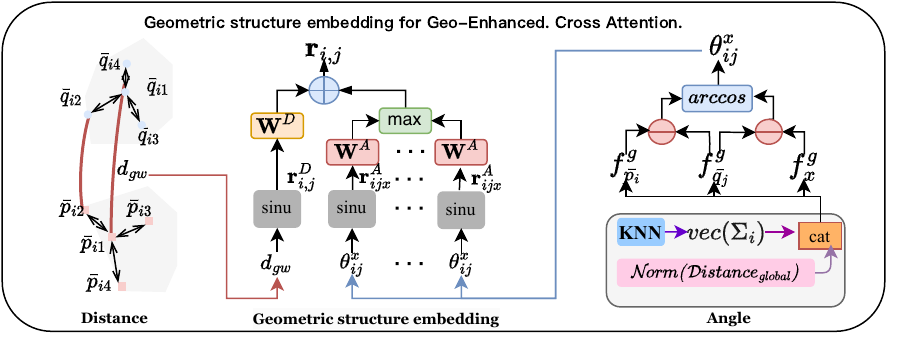}
\caption{
Angle and distance provide geometry information for cross-attention.}
\vspace{-0.4cm}
\label{fig:geo_figure}
\end{figure}

The distance and angle calculated in GeoTransformer are invariant only for rigid transformations within a single point cloud~\cite{qin2022geometric}. 
However, these measures are not invariant to rigid transformations when calculated between different point clouds that live in different reference frames.
To address this issue, we have developed a new way of computing distance and angle measures to compare points between different point clouds.
As shown in Fig.~\ref{fig:geo_figure}, we first uniformly sample $k=10$ neighbors $\Omega_{\bar{p}_i}$ of $\bar{\bm{p}}_i$ in a ball of radius $r>0$, and compute the covariance matrix $\Sigma_{\bar{p}_i}$ of these neighbouring points as:

\begin{equation}
    \begin{aligned}
        \Sigma_{\bar{\bm{p}}_i} &= \sum_{\bm{x}_t\in\Omega_{\bar{\bm{p}}_i}}\omega_{x_t}(\bm{x}_t-\bar{\bm{p}}_i)(\bm{x}_t-\bar{\bm{p}}_i)^\top,\\
        \omega_{x_t} & = \frac{\phi-\|\bm{x}_t-\bar{\bm{p}}_i\|_2}{\sum_{\bm{x}_t\in\Omega_{\bar{\bm{p}}_i}}(\phi-\|\bm{x}_t-\bar{\bm{p}}_i\|_2)}, 
    \end{aligned}
\end{equation}

where $\phi=\max\limits_{\bm{x}_t\in\Omega_{\bar{\bm{p}}_i}}\|\bm{x}_t-\bar{\bm{p}_i}\|_2$.
For $\bar{\bm{p}}_i$, we calculate an eigenvalue tuple of $\Sigma_{\bar{\bm{p}}_i}$ denoted by $\bm{{\lambda}}_{\bar{p}_i}=({\lambda}_{\bar{p}_i}^{1}, {\lambda}_{\bar{p}_i}^{2}, {\lambda}_{\bar{p}_i}^{3})$.
with ${\lambda}_{\bar{p}_i}^{1} \leq {\lambda}_{\bar{p}_i}^{2}\leq {\lambda}_{\bar{p}_i}^{3}$.
The $\Sigma_{\bar{\bm{p}}_i}$ is transformation invariant.
We sign the eigenvector associated with ${\lambda}_{\bar{p}i}^{1}$ as $\bm{v}_{\bar{p}_i}$, ensuring that its orientation is positive from the center of the point cloud towards the current point.
Using the same operation, we get $\bm{{\lambda}}_{\bar{q}_j}$ for $\bar{\bm{q}}_j\in\bar{\bm{\mathcal{Q}}}$ and $\bm{v}_{\bar{q}_j}$.

For distance embedding, we choose the Gromov-Wasserstein distance~\cite{peyre2019computational}, which is transformation invariant, and it can calculate the distance between metrics defined within each of the source and target spaces. 
Therefore, we use the GWD map to measure the transformation relationship $\bm{S}=\{s_{ij}\}$ of points from $\bar{\bm{\mathcal{P}}}$ and $\bar{\bm{\mathcal{Q}}}$, which formulates as:
\begin{equation}\label{eq:gw}
\begin{aligned}
&d_{gw}\left(\bar{\bm{\mathcal{P}}},\bar{\bm{\mathcal{Q}}}\right)=\min_{S\geq0}\sum_{stkl}\left(C_{sk}^{\bar{p}}-{C_{tl}^{\bar{q}}}\right)^2 \bm{S}_{st}\bm{S}_{kl},\\
&\mbox{s.t.,~} 
\begin{cases}
C_{sk}^{\bar{p}}=\|\bar{\bm{p}}_{s}-\bar{\bm{p}}_{k}\|^2_2+\alpha\|\bm{v}_{\bar{p}_s}-\bm{v}_{\bar{p}_k}\|^2_2,\\
C_{sk}^{\bar{q}}=\|\bar{\bm{q}}_{t}-\bar{\bm{q}}_{l}\|^2_2+\alpha\|\bm{v}_{\bar{q}_t}-\bm{v}_{\bar{q}_l}\|^2_2,
\end{cases}
\end{aligned}
\end{equation}
where $\alpha>0$ is a learning parameter. The $\bm{S}_{ij}$ describes the similarity between the points $\bar{\bm{p}}_i$ and $\bar{\bm{q}}_{j}$.
Eq.~\eqref{eq:gw} is solved using the Sinkhorn algorithm~\cite{peyre2019computational}.

For angle computation, we first define $\bm{\lambda}_c=\frac{1}{\bar{N}+\bar{M}}\left(\sum_{i=1}^{\bar{N}}\bm{\lambda}_{\bar{\bm{p}}_i} + \sum_{j=1}^{\bar{M}}\bm{\lambda}_{\bar{\bm{q}}_j}\right)$ as the center of union of $\bar{\bm{\mathcal{P}}}$ and $\bar{\bm{\mathcal{Q}}}$ in the eigenvalue space.
For each $\bar{\bm{p}}_i$, we compute a feature vector
$\bm{f}^g_{\bar{p}_i}= \mbox{cat}\left[\frac{\|\bm{\lambda}_{\bar{\bm{p}}_i}-{\bm{\lambda}}_c\|_2}{\max\limits_j\|\bar{\bm{p}}_j-\bar{\bm{p}}_c\|_2},\bm{\lambda}_{\bar{\bm{p}}_i}\right]$ that encodes the geometric and spatial properties of the local patch of a single point within a single point cloud and the global structure information from two point clouds.
$\mbox{cat}[\cdot, \cdot]$ concatenates two vectors into a single vector.
Applying the same operator to $\bar{\bm{q}}_j\in\bar{\bm{\mathcal{Q}}}$, we can obtain $\bm{f}^g_{\bar{q}_j}$ for $\bar{\bm{q}}_j$.
After that, for each ${\bm{x}\in\Omega_{\bar{\bm{p}}_i}}$, we compute the angle $\theta^x_{ij}$ between the vectors $\bm{f}^g_{\bar{p}_i}-\bm{f}^g_{\bar{q}_j}$ and $\bm{f}^g_{x}-\bm{f}^g_{\bar{q}_j}$ as:
\begin{equation}
    \theta^x_{ij} =\arccos\frac{\left<\bm{f}^g_{\bar{p}_i}-\bm{f}^g_{\bar{q}_j}, \bm{f}^g_{x}-\bm{f}^g_{\bar{q}_j}\right>}{\|\bm{f}^g_{\bar{p}_i}-\bm{f}^g_{\bar{q}_j}\|\| \bm{f}^g_{x}-\bm{f}^g_{\bar{q}_j}\|}.
\end{equation}
The geometric structure embedding of $\bar{\bm{p}}_i\in \bar{\bm{\mathcal{P}}}$ and $\bar{\bm{q}}_j\in \bar{\bm{\mathcal{Q}}}$ comprises a pair-wise cross distance embedding and a triplet-wise cross angular embedding defined as follows:
\begin{itemize}[noitemsep,topsep=0pt,leftmargin=*]
    \item{\verb|Pair-wise Cross Distance Embedding.|} The distance embedding $\bm{r}^D_{ij}$ between $\bar{\bm{p}}_i$ and $\bar{\bm{q}}_j$ is calculated as $\bm{r}^D_{ij}{=}\mbox{sinu}(\left(1-s_{ij}\right) {/} \sigma_{\rho})$, where $\mbox{sinu}(\cdot)$ is a sinusoidal function as in~\cite{vaswani2017attention}, and $\sigma_{\rho}{>}0$ is a learning parameter.
    \item{\verb|Triplet-wise Cross Angle Embedding.|} Angular embedding is computed by using triplets of super-points. 
    Specifically, for each ${\bm{x}\in\Omega_{\bar{\bm{p}}_i}}$, the triplet-wise angular embedding $\bm{r}^A_{ijx}$ is computed as 
    $\bm{r}^A_{ijx}{=}\mbox{sinu}(\theta^x_{ij}/\sigma_{\theta})$ with a learning parameter $\sigma_{\theta}{>}0$.
\end{itemize}
Lastly, the geometric structure embedding $\bm{r}_{ij}$ is obtained by combining the pair-wise distance embedding and the triplet-wise angular embedding through aggregation, expressed as:
\begin{equation}
\bm{r}_{ij} = \bm{r}^D_{ij}\bm{W}^D+\max_x\{\bm{r}^A_{ijx}\bm{W}^A\},
\end{equation}
where  $\bm{W}^D,\bm{W}^A$ are projection matrices corresponding to the two kinds of embeddings, respectively.
The latent features $\hat{\bm{\mathcal{F}}}_{\bar{p}}$ then carry the knowledge of $\hat{\bm{\mathcal{F}}}_{\bar{q}}$, and vice versa. 

\noindent\textbf{Decoder.} 
The decoder, based on KPConv layers \cite{thomas2019kpconv}, starts from the super-points $\bar{\bm{\mathcal{P}}}$ and the concatenations of $\hat{\bm{\mathcal{F}}}_{\bar{p}}$ and  $\bm{\mu}_{\bar{p}}$, and outputs the point cloud $\bm{\mathcal{P}}$ with associated features $\bm{\mathcal{F}}_{p}\in\mathbb{R}^{N\times 32}$ and overlap scores $\bm{\mu}_{p}\in[0,1]^{N}$. The raw point cloud $\bm{\mathcal{Q}}$ and its associated features $\bm{\mathcal{F}}_{q}\in\mathbb{R}^{M\times 32}$ are obtained in the same way.

\noindent\textbf{Local Geometric Self-attention.}
For each super-point, we first construct a local patch of points around it using the point-to-node grouping strategy~\cite{yu2021cofinet}.
For a super-point $\bar{\bm{p}}_i\in \bar{\bm{\mathcal{P}}}$, its associated point set $G_{\bar{p}_i}$ and feature set $G_{\bm{f}_{\bar{p}_i}}$ are denoted as:
\begin{equation}
    \begin{cases}
		G_{\bar{p}_i}=\{\bm{p}\in \bm{\mathcal{P}}\big|\|\bm{p}-\bm{\bar{p}}_i\|_2\leq\|\bm{p}-\bm{\bar{p}}_j\|_2, i\neq j\}, \\
		G_{\bm{f}_{\bar{p}_i}}=\{\bm{f}_{\bm{x}_j}\in \bm{\mathcal{F}}_{p}\big|\bm{x}_j \in G_{\bar{p}_i}\}.
	\end{cases}
\end{equation}
In a similar way, we can get $G_{\bar{q}_i}$ and $G_{\bm{f}_{\bar{q}_i}}$.

Given the local patch $G_{\bar{p}_i}=\{\bar{\bm{p}}_{i1},\bar{\bm{p}}_{i2}, \cdots,\bar{\bm{p}}_{iK}\}$, 
to perform local geometric attention, we first calculate the distance matrix $D^i=\{d^i_{kl}\}_{k,l=1}^K$, 
where $d^i_{kl}=\{\bar{\bm{p}}_{i1},\bar{\bm{p}}_{i2}, \cdots, \bar{\bm{p}}_{iK}\}$ with $d^i_{kl}=\|\bar{\bm{p}}_{ik}-\bar{\bm{p}}_{il}\|^2_2$.
Distance-based weights are calculated by $R^i=\mathtt{DS}(D^i)=\{r^i_{kl}\}$. $\mathtt{DS}(\dot)$ is a Dual Softmax operator.
Then, the self-attention output $\bm{\bm{f}}^i_{\bar{p}_{ik}}\in G_{\bm{f}_{\bar{p}_i}}$ for $\bm{\bar{p}}_{ik}$ can be updated by computing the weighted sum of all weighted input features:
\vspace{-0.2cm}
\begin{equation}
    \bm{f}^i_{\bar{p}_{ik}} = \sum_{l=1}^{K}\frac{\beta_{kl}}{\sum_j\beta_{kj}}\bm{f}^i_{\bar{p}_{il}},
\end{equation}
where the weight coefficient $\beta_{kl}$ is calculated using row-wise softmax on the attention scores as:
\begin{equation}
    \beta_{kl} = r_{kl}^i\cdot\mbox{softmax} \left(  \frac{\bm{f}^i_{\bar{p}_{ik}}\bm{W}_i^Q(\bm{f}^i_{\bar{p}_{il}}\bm{W}_i^K)^\top}{\sqrt{b}}\right),
\end{equation}
We also applied the same operator to update $G_{\bm{f}_{\bar{q}_i}}$.

\subsubsection{Correspondence Prediction}

\noindent \textbf{Coarse Matching.}
Coarse Matching estimates super-point correspondences between $\bm{\bar{\mathcal{P}}}$ and $\bm{\bar{\mathcal{Q}}}$, which can be formulated as an assignment problem and solved by calculating an assignment matrix $\bar{\Gamma}\in \mathbb{R}^{\bar{N}\times \bar{M}}$ as:
\begin{equation}\label{eq:super}
\min_{\bar{\Gamma}}\left<\bar{\bm{C}}, \bar{\Gamma}\right>, 
\end{equation}
where $\left<\cdot, \cdot\right> $ is the Frobenius dot product. 
$\bar{\bm{C}}$ is a distance matrix with elements that satisfy 
\begin{small}
\begin{equation*}
\bar{\bm{C}}_{ij}=(1-\eta)\|\frac{\hat{\bm{f}}_{\bar{p}_i}}{\|\hat{\bm{f}}_{\bar{p}_i}\|}-\frac{\hat{\bm{f}}_{\bar{p}_j}}
{\|\hat{\bm{f}}_{\bar{p}_j}\|}\|
+
\eta\|\frac{\bm{f}^g_{\bar{p}_i}}{\|\bm{f}^g_{\bar{p}_i}\|}-\frac{{\bm{f}}^g_{\bar{q}_j}}{\|\bm{f}^g_{\bar{q}_j}\|}\|,
\end{equation*}
\end{small}
with  $\hat{\bm{f}}_{\bar{p}_i}\in\hat{\bm{\mathcal{F}}}_{},\hat{\bm{f}}_{\bar{q}_j}\in\hat{\bm{\mathcal{F}}}_{\bar{q}}$, and each $\bar{\Gamma}_{ij} \in \bar{\Gamma}\in[0,1]^{\bar{N}\times \bar{M}}$ represents the matching score between $\bm{\bar{p}}_i$ and $\bm{\bar{q}}_j$.
$\eta=0.1$ is a parameter that controls the weight of the feature distance and the structure distance.
Eq.~\eqref{eq:super} is an example of the optimal transport problem \cite{cuturi2013sinkhorn} and can be solved efficiently using the Sinkhorn-Knopp algorithm \cite{cuturi2013sinkhorn}.
After computing $\bar{\Gamma}$, we select correspondences with confidence higher than a threshold $\tau_c$, and enforce the mutual nearest neighbor (MNN) constraint to have fewer but reliable correspondences. 
The super-point correspondence set $\mathcal{\bar{M}}$ is then defined as:
\begin{equation}\label{eq:sset}
	\mathcal{\bar{M}}=\{(\bm{\bar{p}}_{\hat{i}},\bm{\bar{q}}_{\hat{j}})\big|\forall(\hat{i},\hat{j})\in \mbox{MNN}(\bar{\Gamma}), \bar{\Gamma}_{\hat{i},\hat{j}}\geq\tau_c\}.
\end{equation}

\noindent \textbf{Fine Matching.}
Extracting point correspondences is analogous to matching two smaller-scale point clouds by solving an optimal transport problem to calculate a matrix $\Gamma^{G_{\bar{p}_i}}$ in a manner of coarse-level done.
For correspondences, we choose the maximum confidence score of $\Gamma^{G_{\bar{p}_i}}$ in every row and column to guarantee higher precision. The final point correspondence set $\mathcal{M}$ is represented as the union of all the correspondence sets obtained. After obtaining the correspondences $\mathcal{M}$, following \cite{qin2022geometric,yu2021cofinet},  a variant of RANSAC \cite{fischler1981random} that is specialized in 3D correspondence-based registration \cite{zhou2018open3d} is utilized to estimate the transformation.

\subsection{Loss Function and Training}
We train \ourmethod using ground-truth correspondences as supervision. 
The loss function is: $\mathcal{L} = \mathcal{L}_{C}  + \mathcal{L}_{F}$,
where $\mathcal{L}_{C}$ and $\mathcal{L}_{F}$ denote a coarse and a fine matching loss.

\noindent\textbf{Coarse Matching Loss.}
Following \cite{yu2021cofinet,fu2021robust}, we formulate super-point matching as a multilabel classification problem and adopt a cross-entropy loss with optimal transport. As there is no direct supervision for super-point matching, we leverage the overlap ratio $r_{ij}$ of points in $G_{\bar{p}_i}$ that have correspondences in $G_{\bar{q}_j}$ to depict the matching probability between super-points $\bar{p}_i$ and $\bar{q}_j$. $r_{ij}$ is defined as:
\begin{equation}\label{eq:cratio}
r_{ij} =\frac{1}{{|G_{\bar{p}_i}|}} {|\{\bm{p}\in G_{\bar{p}_i}\big|\min_{\bm{q}\in G_{\bar{q}_j}}\|\hat{\bm{T}}\left(\bm{p}\right) -\bm{q}\|_2<r_p\}|},
\end{equation}
where $\hat{\bm{T}}$ is the ground-truth transformation and $r_p$ is a set threshold.
All other pairs are omitted. We select the super-points in $\bm{\bar{\mathcal{P}}}$ which have at least one positive super-point in $\bm{\bar{\mathcal{Q}}}$ to form a set of anchor super-points, $\bm{\tilde{\mathcal{P}}}$. 
Based on Eq.~\eqref{eq:cratio}, we define the weight matrix $\mathbf{\bar{W}} \in \mathbb{R}^{\bar{N} \times \bar{M}}$ as:
\begin{scriptsize}
\begin{equation*}
    \mathbf{\bar{W}}\left(i, j\right){=} 
    \begin{cases}
    r\left(i, j\right),  & i \leq \bar{N} \wedge j \leq \bar{M}, \\  
    0, & \text {otherwise.}
    \end{cases}
\end{equation*}
\end{scriptsize}
Finally, the coarse matching loss can be written as:
\begin{equation}
\mathcal{L}_C=-\frac{\sum_{i, j} \mathbf{\bar{W}}\left(i, j\right) \log \left(\mathbf{\bar{\Gamma}}\left(i, j\right)\right)}{\sum_{i, j} \mathbf{\bar{W}}\left(i, j\right)}.
\end{equation}

\noindent\textbf{Fine Matching Loss.}
We apply the circle loss again to supervise the point matching.
Consider a pair of matched super-points $\bar{\bm{p}}_i$ and $\bar{\bm{q}}_j$ with associated patches $G_{\bar{p}_i}$ and $G_{\bar{q}_j}$, we first extract a set of anchor points $\tilde{G}_{\bar{p}_i} \subseteq G_{\bar{p}_i}$ satisfying that each $\bm{g}^k_{\bar{p}_i}\in \tilde{G}_{\bar{p}_i}$ has at least one (possibly multiple) correspondence in $G_{\bar{q}_j}$, i.e.,
\begin{equation*}
\tilde{G}_{\bar{p}_i} = \{\bm{g}^k_{\bar{p}_i}\in \tilde{G}_{\bar{p}_i} |\min_{\bm{g}^l_{\bar{q}_j}\in G_{\bar{q}_j}}\|\hat{\bm{T}}\left(\bm{g}^k_{\bar{p}_i}\right)-\bm{g}^l_{\bar{q}_j}\|_2 < r_p\}.
\end{equation*}
For each anchor $\bm{g}^k_{\bar{p}_i}\in \tilde{G}_{\bar{p}_i}$,  we denote the set of its positive points in $G_{\bar{q}_j}$ as $\mathcal{N}_p^{\bm{g}^k_{\bar{p}_i}}$. All points of $\mathcal{\bm{Q}}$ outside a (larger) radio $r_n$ form the set of its negative patches as $\mathcal{N}_n^{\bm{g}^k_{\bar{p}_i}}$. The fine-level matching loss $\mathcal{L}_{F}^{\mathcal{\bm{P}}}$ on $\mathcal{\bm{P}}$ is calculated as:
\vspace{-2mm}
\begin{scriptsize}
\begin{equation*}
\begin{aligned}
    \mathcal{L}_{F}^{\mathcal{\bm{P}}} &=\frac{1}{|\bm{\tilde{\mathcal{P}}}|}\sum_{\tilde{\bm{p}}_i\in\bm{\bar{\mathcal{P}}}}\frac{1}{|\tilde{G}_{\bar{p}_i}|} \sum_{\bm{g}^s_{\bar{p}_i}\in \tilde{G}_{\bar{p}_i}}\log\left[1+\xi_s\right], \\
    \xi_s &= \sum_{\bm{g}^k_{\bar{q}_j}\in\mathcal{N}_p^{\bm{g}^s_{\bar{p}_i}}}e^{r^k_s\beta_p^{sk}(d_s^k-\Delta p)}\cdot\sum_{\bm{g}^l_{\bar{q}_j}\in\mathcal{N}_n^{\bm{g}^s_{\bar{p}_i}}}e^{\beta_n^{sl}(\Delta n - d_s^l)},
\end{aligned}
\end{equation*}
\end{scriptsize}
where $d_s^k=\mathcal{D}_f(\bm{f}_{\bm{g}^s_{\bar{p}_i}},\bm{f}_{\bm{g}^s_{\bar{q}_j}})$ is the distance in the feature space. The weights $\beta_p^{sk}=\omega d_s^k$ and $\beta_n^{sl}{=}\omega (2.0-d_s^l)$ are determined individually for each positive and negative example with a learned scale factor $\omega\geq 1$. $\Delta p = 0.1$ and $\Delta n = 1.4$. The same goes for the loss $\mathcal{L}_{F}^{\bm{\mathcal{Q}}}$ on $\bm{\mathcal{Q}}$. The
overall super-point matching loss is written as
$\mathcal{L}_{F} {=} \frac{1}{2}(\mathcal{L}_{F}^{\mathcal{\bm{P}}} + \mathcal{L}_{F}^{\mathcal{\bm{Q}}})$.
\section{Experiments}
\label{sec:exp}
We evaluate \ourmethod on typical point cloud registration benchmarks, i.e.~indoor 3DMatch \cite{zeng20173dmatch} and 3DLoMatch \cite{huang2021predator}, outdoor KITTI \cite{geiger2012we}, and cross-source 3DCSR \cite{huang2021comprehensive}.
Please refer to Appendix Section 1 for detailed implementation, including running details, network pipeline, and correspondence sampling.
For cluster numbers and the time computational analysis, and additional qualitative results please refer to the Sec.~2 in the Supplementary Material.

\subsection{Evaluation on 3DMatch and 3DLoMatch}

\noindent\textbf{Datasets.} 
3DMatch \cite{zeng20173dmatch} and 3DLoMatch \cite{huang2021predator} are two widely used indoor benchmarks that contain more than $30\%$ and 10\% to 30\% partially overlapping scene pairs, respectively. 
3DMatch contains 62 scenes: we use 46 scenes for training, 8 scenes for validation, and 8 scenes for testing. 
The test set contains 1,623 partially overlapped point cloud fragments and their corresponding transformation matrices. 
We use training data preprocessed by \cite{huang2021predator} and evaluate on both the 3DMatch and 3DLoMatch \cite{huang2021predator} protocols. 
We first voxelize the point clouds with a $2.5cm$ voxel size and then extract different feature descriptors. 
We set $\tau_c=0.15$, $r=r_o=r_p=3.75cm$, and $r_n=10.0cm$ \cite{huang2021predator}.

\noindent\textbf{Metrics.} 
Following Predator \cite{huang2021predator} and CoFiNet \cite{yu2021cofinet}, we evaluate performance with three metrics: 
(i) \textit{Inlier Ratio} (IR), the fraction of putative correspondences whose residuals are below a certain threshold (i.e.~0.1m) under the ground-truth transformation, 
(ii) \textit{Feature Matching Recall} (FMR), the fraction of point cloud pairs whose inlier ratio is above a certain threshold (i.e.~5\%), and 
(iii) \textit{Registration Recall} (RR), the fraction of point cloud pairs whose transformation error is smaller than a certain threshold (i.e.~$RMSE<0.2m$). 
We compare \ourmethod with FCGF \cite{choy2019fully}, D3Feat \cite{bai2020d3feat}, SpinNet \cite{ao2021spinnet}, Predator \cite{huang2021predator}, YOHO \cite{wang2022you}, CoFiNet \cite{yu2021cofinet}, GeoTransformer \cite{qin2022geometric} short as GeoTrans, GLORN \cite{xu2022glorn}, and RoITr~\cite{yu2023rotation}.

\begin{table}[t]
\centering
	\caption{Results on both 3DMatch and 3DLoMatch datasets under different numbers of samples. Best performance in bold.}
 \resizebox{0.95\linewidth}{!}{%
	\setlength{\tabcolsep}{0.5mm}
	{
		\begin{tabular}{l | c c c c c | c c c c c}
			\toprule
			~ &  \multicolumn{5}{c|}{3DMatch} &  \multicolumn{5}{c}{3DLoMatch} \\
			\# Samples  & 5000 & 2500 & 1000 & 500 & 250 & 5000 & 2500 & 1000 & 500 & 250 \\
			\midrule
			Method & \multicolumn{10}{c}{Inlier Ratio $(\%) \uparrow$} \\
			\hline
			FCGF\cite{choy2019fully}  		& 56.8 & 54.1 & 48.7 & 42.5 & 34.1 & 21.4 & 20.0 & 17.2 & 14.8 & 11.6\\
			D3Feat\cite{bai2020d3feat}  	& 39.0 & 38.8 & 40.4 & 41.5 & 41.8 & 13.2 & 13.1 & 14.0 & 14.6 & 15.0\\
			SpinNet \cite{ao2021spinnet} 	& 47.5 & 44.7 & 39.4 & 33.9 & 27.6 & 20.5 & 19.0 & 16.3 & 13.8 & 11.1 \\
			Predator \cite{huang2021predator}  & 58.0 & 58.4 & 57.1 & 54.1 & 49.3 & 26.7 & 28.1 & 28.3 & 27.5 & 25.8\\
			CoFiNet\cite{yu2021cofinet}		& 49.8 & 51.2 & 51.9 & 52.2 & 52.2 & 24.4 & 25.9 & 26.7 & 26.8 & 26.9\\
			YOHO \cite{wang2022you} 	& 64.4 & 60.7 & 55.7 & 46.4 & 41.2 & 25.9 & 23.3 & 22.6 & 18.2 & 15.0\\
			GeoTrans\cite{qin2022geometric} & 71.9 & 75.2 & 76.0 & 82.2 & 85.1 
			& 43.5 & 45.3 & 46.2 & 52.9 & 57.7 \\
                GLORN \cite{xu2022glorn} & 72.6 & 73.5 & 76.4 & 82.3 & 83.2 &
                42.3 & 45.6 & 46.5 & 53.1 & \bf57.9 \\
                RoITr~\cite{yu2023rotation} & 82.6 & 82.8 & 83.0 & 83.0 & 83.0 & 54.3 & 54.6 & 55.1 & 55.2 & 55.3 \\
			\ourmethod(Ours)	& \bf83.1 & \bf83.6 & \bf84.2 & \bf84.2 & \bf84.1 & \bf56.1 & \bf56.4 & \bf57.3 & \bf57.3 & 57.4\\
			\hline
			& \multicolumn{10}{c}{Feature Matching Recall $(\%) \uparrow$} \\
			\hline
			FCGF\cite{choy2019fully} 		& 97.4 & 97.3 & 97.0 & 96.7 & 96.6 & 76.6 & 75.4 & 74.2 & 71.7 & 67.3 \\
			D3Feat \cite{bai2020d3feat}  	& 95.6 & 95.4 & 94.5 & 94.1 & 93.1 & 67.3 & 66.7 & 67.0 & 66.7 & 66.5 \\
			SpinNet \cite{ao2021spinnet} 	& 97.6 & 97.2 & 96.8 & 95.5 & 94.3 & 75.3 & 74.9 & 72.5 & 70.0 & 63.6 \\
			Predator\cite{huang2021predator} & 96.6 & 96.6 & 96.5 & 96.3 & 96.5 & 78.6 & 77.4 & 76.3 & 75.7 & 75.3\\
			CoFiNet\cite{yu2021cofinet}	& 98.1 & 98.3 & 98.1 & \bf98.2 & 98.3 & 83.1 & 83.5 & 83.3 & 83.1 & 82.6\\
			YOHO\cite{wang2022you} 		& 98.2 & 97.6 & 97.5 & 97.7 & 96.0 & 79.4 & 78.1 & 76.3 & 73.8 & 69.1\\
			GeoTrans\cite{qin2022geometric} & 97.9 & 97.9 & 97.9 & 97.9 & 97.6 & 88.3 & 88.6 & 88.8 & 88.6 & 88.3 \\
                GLORN \cite{xu2022glorn} & 97.8 & 97.9 & 98.0 & 98.1 & 97.2 & 87.8 & 88.8 & 88.9 & 88.7 & 88.5 \\
                RoITr~\cite{yu2023rotation} & 98.0 & 98.0 & 97.9 & 98.0 & 97.9 & \bf89.6 & \bf89.6 & 89.5 & 89.4 & 89.3 \\
			\ourmethod(Ours)  & \bf98.2 & \bf98.2 & \bf98.1 & \bf98.2 & \bf98.3 & 89.5 & \bf89.6 & \bf89.6 & \bf89.5 & \bf89.4 \\
			\hline
			& \multicolumn{10}{c}{Registration Recall $(\%) \uparrow$} \\
			\hline
			FCGF\cite{choy2019fully} 	& 85.1 & 84.7 & 83.3 & 81.6 & 71.4 & 40.1 & 41.7 & 38.2 & 35.4 & 26.8 \\
			D3Feat\cite{bai2020d3feat} 	& 81.6 & 84.5 & 83.4 & 82.4 & 77.9 & 37.2 & 42.7 & 46.9 & 43.8 & 39.1 \\
			SpinNet\cite{ao2021spinnet} & 88.6 & 86.6 & 85.5 & 83.5 & 70.2 & 59.8 & 54.9 & 48.3 & 39.8 & 26.8 \\
			Predator\cite{huang2021predator} 	& 89.0 & 89.9 & 90.6 & 88.5 & 86.6 & 59.8 & 61.2 & 62.4 & 60.8 & 58.1 \\
			CoFiNet\cite{yu2021cofinet}	  & 89.3 & 88.9 & 88.4 & 87.4 & 87.0 & 67.5 & 66.2 & 64.2 & 63.1 & 61.0 \\
			YOHO \cite{wang2022you} 	  & 90.8 & 90.3 & 89.1 & 88.6 & 84.5 & 65.2 & 65.5 & 63.2 & 56.5 & 48.0 \\
			GeoTrans\cite{qin2022geometric} & 92.0 & 91.8 & 91.8 & 91.4 & 91.2 & 75.0 & 74.8 & 74.2 & 74.1 & 73.5 \\
            GLORN \cite{xu2022glorn} & 92.2 & 91.6 & 91.9 & 91.5 & 91.0 & 74.9 & 75.2 & 74.4 & 74.3 & 73.6 \\
            RoITr~\cite{yu2023rotation} & 91.9 & 91.7 & 91.8 & 91.4 & 91.0 & 74.7 & 74.8 & 74.8 & 74.2 & 73.6 \\
			\ourmethod(Ours) & \bf92.4 & \bf92.1 & \bf92.2 & \bf91.8 & \bf91.5 & \bf78.6 & \bf78.7 & \bf78.1 & \bf76.4 & \bf75.2 \\
			\bottomrule
		\end{tabular}
	}}
	\label{table:3dm}
 \vspace{-0.4cm}
\end{table}

\noindent\textbf{Inlier Ratio and Feature Matching Recall.} 
The primary contribution of our \ourmethod is its use of geometric cross-attention to emphasize matched point pair similarities and estimate more accurate correspondences. Therefore, we begin by examining the inlier ratio of the correspondences generated by \ourmethod, which directly reflects the quality of the extracted correspondences.
Following \cite{huang2021predator}, we present the results on varying sampled numbers of correspondences. 
Table~\ref{table:3dm} (top) shows that \ourmethod outperforms all previous methods in terms of Inlier Ratio on both benchmarks.
In particular, \ourmethod consistently outperforms RoITr, the second best baseline, by $0.5\%\sim 1.2\%$ on 3DMatch and $1.8\%\sim 2.1\%$ on 3DLoMatch, with sample numbers ranging from 250 to 5000. 
This notable increase in the Inlier Ratio suggests that incorporating the cross-geometric structure effectively enhances the reliability of correspondence production. Additionally, \ourmethod outshines all competitors in Feature Matching Recall, as detailed in Table~\ref{table:3dm} (middle section). Particularly in the challenging low-overlap scenarios of 3DLoMatch, our method demonstrates its robustness with improvements exceeding 0.4\%, underscoring its efficacy in complex situations.

\begin{figure}[t]
\centering 
\begin{overpic}[width=0.85\columnwidth]{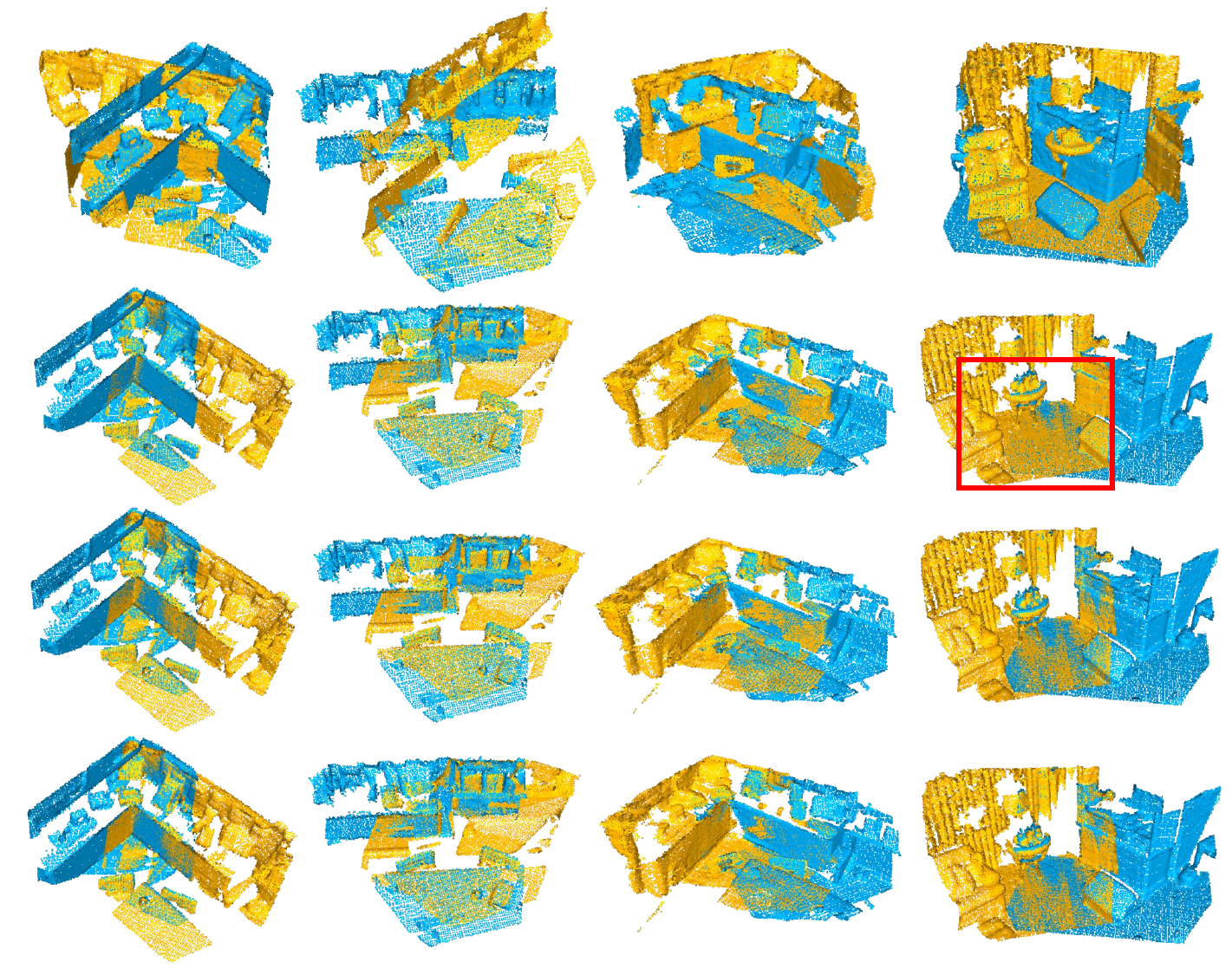}
    \put(-3.7,65.0){\color{black}\footnotesize\rotatebox{90}{\textbf{Input}}}
    \put(-3.7,42.0){\color{black}\footnotesize\rotatebox{90}{\textbf{GeoTrans}}}
    \put(-3.7,25.4){\color{black}\footnotesize\rotatebox{90}{\textbf{Ours}}}
    \put(-3.7,7.0){\color{black}\footnotesize\rotatebox{90}{\textbf{GT}}}
\end{overpic}
\caption{Examples of qualitative registration results on the 3DMatch dataset. 
Inaccurate regions are enclosed in red boxes.}
	\label{fig:3dvs}
\end{figure}

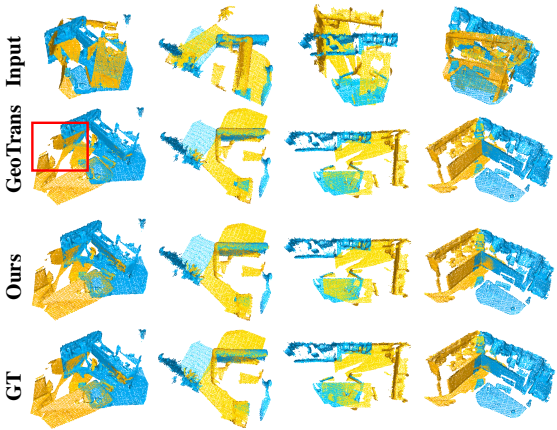
\begin{figure}[t]
	\centering 
\begin{overpic}[width=0.85\columnwidth]{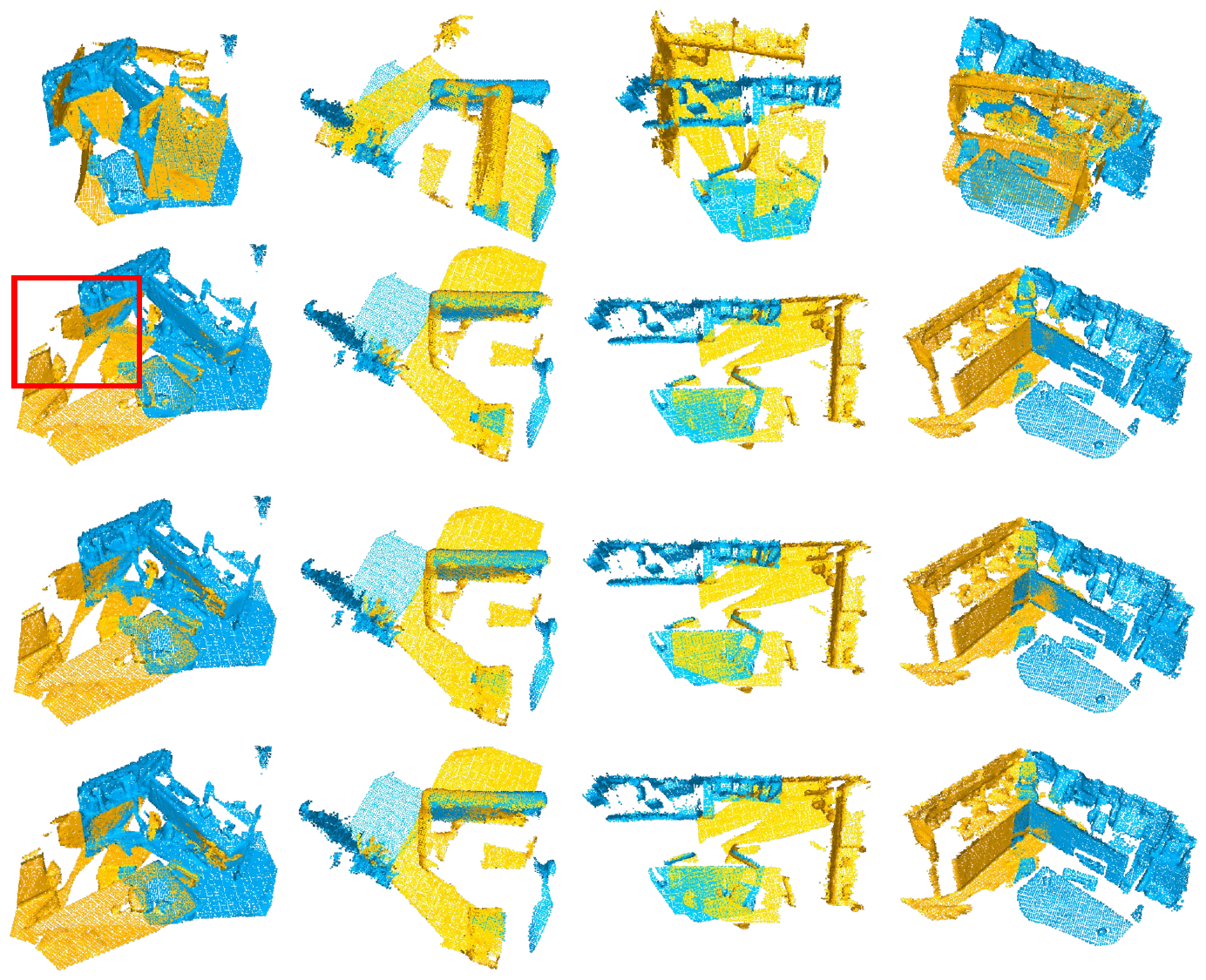}
    \put(-3.7,65.0){\color{black}\footnotesize\rotatebox{90}{\textbf{Input}}}
    \put(-3.7,45.0){\color{black}\footnotesize\rotatebox{90}{\textbf{GeoTrans}}}
    \put(-3.7,25.0){\color{black}\footnotesize\rotatebox{90}{\textbf{Ours}}}
    \put(-3.7,6.0){\color{black}\footnotesize\rotatebox{90}{\textbf{GT}}}
\end{overpic}
\caption{Examples of qualitative registration results on the 3DLoMatch dataset.
Inaccurate regions are enclosed in red boxes.}
\label{fig:3dvslo}
 \vspace{-0.4cm}
\end{figure}

\noindent\textbf{Registration Recall.} 
The Registration Recall (RR) reflects the final performance on point cloud registration. 
Table~\ref{table:3dm} (bottom) shows that our method outperforms all other models on both datasets in terms of RR. 
\ourmethod achieves RR of $92.4\%$ and $78.7\%$ on both 3DMatch and 3DLoMatch, surpassing the previous best performance achieved by GeoTransformer (which has a RR of $92.0\%$ on 3DMatch and $75.0\%$ on 3DLoMatch) by $0.4\%$ and $3.7\%$, respectively. 
This shows that incorporating geometrical information into the correspondence prediction process can alleviate ambiguity and lead to superior performance compared to methods that only consider feature similarity in cross-attention.
Figs.~\ref{fig:3dvs} and \ref{fig:3dvslo} show comparison examples on 3DMatch and 3DLoMatch, respectively. 
\ourmethod achieves better results in challenging indoor scenes with a low overlap ratio.

\subsection{Evaluation on KITTI}

\noindent\textbf{Dataset.} 
KITTI consists of 11 sequences of LiDAR-scanned outdoor driving scenarios. To ensure fairness in comparisons, we adopt the same data-splitting approach as \cite{choy2019fully, choy2020deep}, with sequences 0-5 utilized for training, 6-7 for validation, and 8-10 for testing purposes. 
We refine the provided ground-truth poses by employing ICP and limit the evaluation to point cloud pairs within a distance of $10m$ from each other, as in \cite{choy2020deep}. 
Furthermore, following in \cite{huang2021predator}, we downsample the point clouds using a voxel size $30cm$ and set $\tau_c=0.15$, $r=r_o$=$45cm$, $r_p$=$21cm$, and $r_n$=$75cm$.

\noindent\textbf{Metrics.} 
Following Predator~\cite{huang2021predator} and CoFiNet~\cite{yu2021cofinet}, our \ourmethod is evaluated using three metrics: Registration Recall (RR), Relative Rotation Error (RRE), and Relative Translation Error (RTE). RR calculates the percentage of successful alignments where the rotation and translation errors are below specified thresholds (i.e., RRE $<5^\circ$ and RTE $< 2m$). The definitions of RRE and RTE are RRE$=\arccos\frac{\textbf{Tr}\left(\bm{R}^\top\bm{R}^\star\right)-1}{2}$ and RTE$=|\bm{t}-\bm{t}^\star|_2$, respectively. The ground-truth rotation matrix and the translation vector are denoted by $\bm{R}^\star$ and $\bm{t}^\star$.

\noindent\textbf{Registration Results.} 
We compare \ourmethod against FCGF \cite{choy2019fully}, D3Feat \cite{bai2020d3feat}, SpinNet \cite{ao2021spinnet}, Predator \cite{huang2021predator}, CoFiNet \cite{yu2021cofinet}, and GeoTransformer \cite{qin2022geometric}. 
Table~\ref{table:kitti} shows that our method achieves the best performance in terms of RR and the lowest average RTE and RRE.
These results confirm the effectiveness of \ourmethod also in an outdoor scenario.
It also suggests that incorporating cross-attention with geometry enhancement could be beneficial in acquiring more distinct features, resulting in better performance in registration.



\renewcommand{\arraystretch}{0.85}
\begin{table}[t]
	\centering
    \small
	\caption{Results on KITTI dataset. Best performance in bold.}
    \resizebox{0.8\linewidth}{!}{%
	\begin{tabular}{l | c | c c c}
		\toprule
		Method & RTE (cm) $\downarrow$ & RRE $(^\circ)\downarrow$ & RR(\%) $\uparrow$ \\
		\midrule
		FCGF \cite{choy2019fully} 	& 9.5 & 0.30 & 96.6 \\
		D3Feat \cite{bai2020d3feat} 	 & 7.2 & 0.30 & \bf99.8 \\
		SpinNet \cite{ao2021spinnet} & 9.9 & 0.47 & 99.1 \\
		Predator \cite{huang2021predator}  & 6.8 & 0.27 & \bf99.8 \\	
		CoFiNet \cite{yu2021cofinet}	   & 8.5 & 0.41 & \bf99.8 \\
		GeoTrans \cite{qin2022geometric} & 7.4 & 0.27 & \bf99.8 \\
		\ourmethod (ours)					& \bf6.9 & \bf0.24 & \bf99.8 \\
		\bottomrule
	\end{tabular}
 }
\label{table:kitti}
\vspace{-3mm}
\end{table}
\renewcommand{\arraystretch}{1} 

\renewcommand{\arraystretch}{0.85} 
\begin{table}[t]
\centering
\small
\caption{Registration results on the 3DCSR dataset. Best performance in bold.}
\resizebox{0.8\linewidth}{!}{%
\begin{tabular}{l | c c c}
    \toprule
    Method & RTE (cm) $\downarrow$ & RRE $(^\circ)\downarrow$ & RR(\%) $\uparrow$ \\
    \midrule
    FCGF \cite{choy2019fully}	 & \bf0.21 & 7.47 & 49.6  \\
    D3Feat \cite{bai2020d3feat} 	 & 0.26 & 6.41 & 52.0 \\
    SpinNet \cite{ao2021spinnet} & 0.24 & 6.56 & 53.5\\
    Predator \cite{huang2021predator}  & 0.27 & 6.26 & 54.6 \\	
    CoFiNet \cite{yu2021cofinet}	   & 0.26 & 5.76 & 57.3 \\
    GeoTrans \cite{qin2022geometric} & 0.24 & 5.60 & 60.2 \\
    \ourmethod (ours)				& 0.22 & \bf5.44 & \bf62.9 \\
    \bottomrule
\end{tabular}
}
\label{table:3DCSR}
\vspace{-5mm}
\end{table}
\renewcommand{\arraystretch}{1} 

\subsection{Generalization on 3D Cross-Source Dataset}

\noindent\textbf{Dataset.}
3DCSR~\cite{huang2021comprehensive} contains two sets: Kinect Lidar and Kinect SFM. 
Kinect Lidar comprises 19 scenes captured from both Kinect and Lidar sensors. 
Each scene is divided into different parts. 
Kinect SFM comprises 2 scenes captured from both Kinect and RGB-D sensors. 
 RGB-D images are transformed into a point cloud by employing the VSFM software.
The model trained on 3DMatch is used since the cross-source dataset is captured in an indoor setting. 
The metric used for successful alignment is RR, which represents the percentage of aligned scenes with RRE less than $15^\circ$ and RTE less than $6m$.
3DCSR is a challenging dataset for registration due to a mixture of noise, outliers, density differences, and partial overlaps.

\begin{figure}[t]
\centering 
\begin{overpic}[width=0.8\columnwidth]{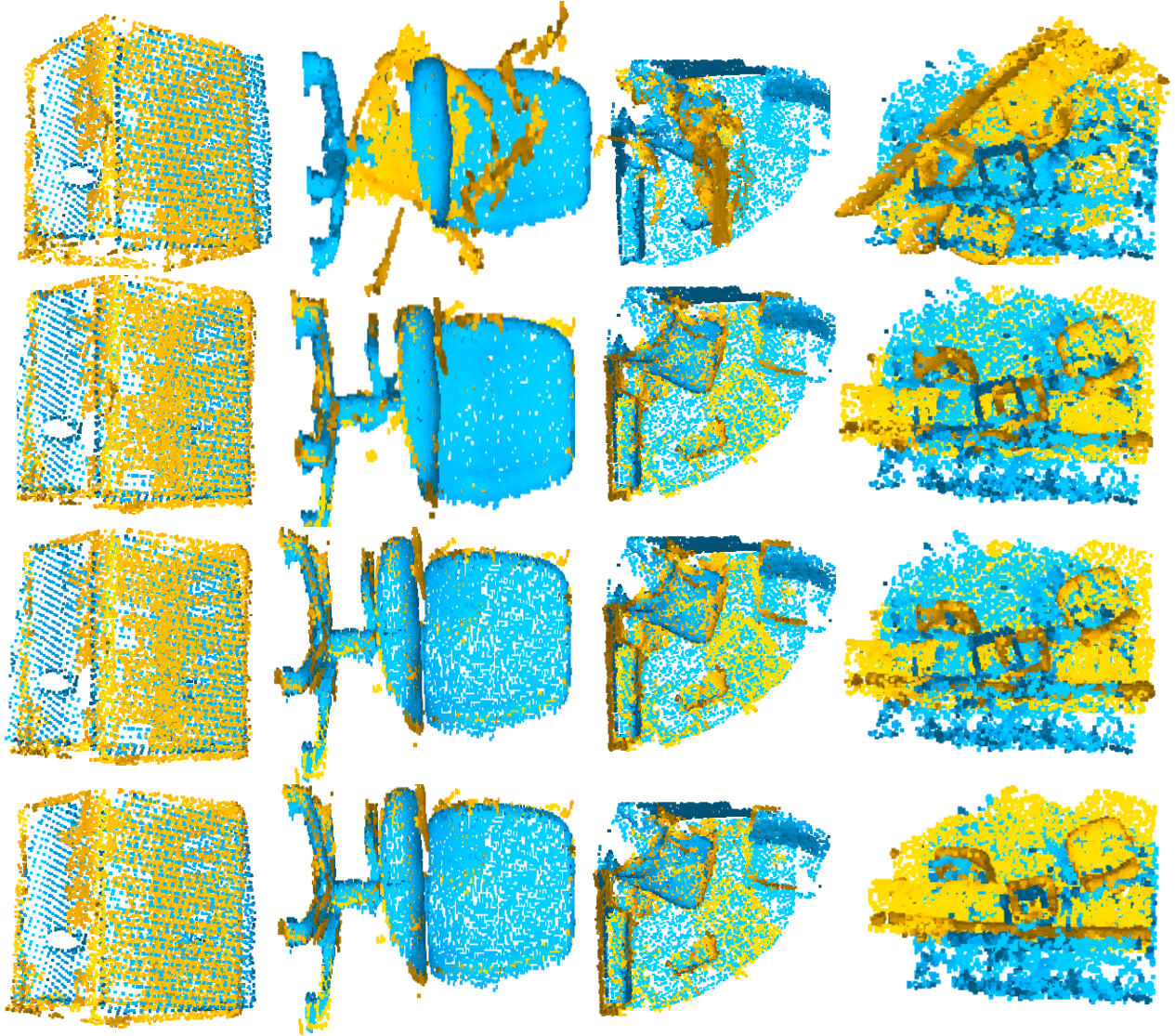}
    \put(-3.7,70.0){\color{black}\footnotesize\rotatebox{90}{\textbf{Input}}}
    \put(-3.7,42.0){\color{black}\footnotesize\rotatebox{90}{\textbf{GeoTrans}}}
    \put(-3.7,25.4){\color{black}\footnotesize\rotatebox{90}{\textbf{Ours}}}
    \put(-3.7,7.0){\color{black}\footnotesize\rotatebox{90}{\textbf{GT}}}
\end{overpic}
\caption{Examples of qualitative registration results on the 3DCSR dataset.}
\label{fig:csdvs}
\end{figure}
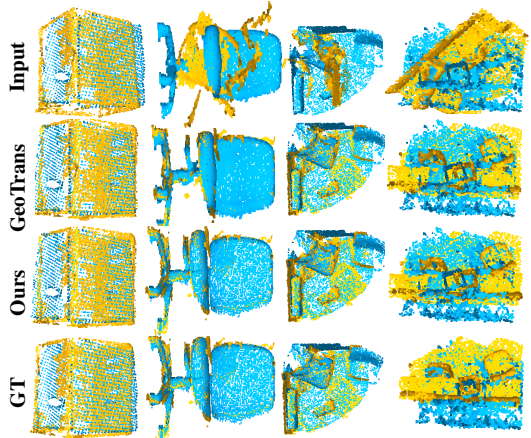

\noindent\textbf{Registration Results.} 
We use FCGF \cite{choy2019fully}, D3Feat \cite{bai2020d3feat}, SpinNet \cite{ao2021spinnet}, Predator \cite{huang2021predator}, CoFiNet \cite{yu2021cofinet}, and GeoTransformer \cite{qin2022geometric} as comparison methods. 
Table~\ref{table:3DCSR} shows that our method also achieves the highest accuracy in this experimental configuration, i.e.~generalization ability. 
Notably, it surpasses GeoTransformer, the second-best, by more than 2.7\% in terms of registration recall (62.9\% vs 60.2\%).
Our ability to achieve better results is attributed to the cross-attention module enhanced by geometry.
However, the recall rate falls short, indicating that the registration challenges on 3DCSR persist.
Fig.~\ref{fig:csdvs} shows examples of qualitative results on 3DCSR.

\begin{table}[t]
    \renewcommand{\arraystretch}{0.8} 
	\centering
	\caption{Ablation study of model design.} 
 \resizebox{0.85\linewidth}{!}{%
	\begin{tabular}{c | c c c | c c c}
	\toprule
	& \multicolumn{3}{c}{3DMatch} & \multicolumn{3}{c}{3DLoMatch} \\
	\hline
        Model & RR & FMR & IR & RR & FMR & IR \\
        \midrule
        \multicolumn{7}{c}{Cross-attention} \\
	\midrule
	vanilla cross-attention & 92.0 & 98.0 & 73.4 & 74.3 & 88.6 & 43.8 \\
	cross-attention w/PDE   & 92.1 & 98.1 & 81.4 & 75.6 & 88.7 & 54.4 \\ 
	cross-attention w/TAE   & 92.1 & 98.0 & 81.3 & 74.9 & 88.6 & 54.1\\
        cross-attention w/DAE   & \bf92.2 & \bf98.1 & \bf84.2 & \bf78.1 & \bf89.6 & \bf57.3 \\
            \midrule
        \multicolumn{7}{c}{Self-attention} \\
        \midrule
        vanilla self-attention & 92.1 & 98.0 & 83.9 & 75.1 & 88.8 & 56.4 \\
	self-attention w/PDE   & \bf92.2 & \bf98.1 & \bf84.2 & \bf78.1 & \bf89.6 & \bf57.3 \\ 
	\bottomrule
	\end{tabular}
 }
	\label{tb:abcross}
\vspace{-5mm}
\end{table}

\vspace{-1mm}

\subsection{Ablation Study}
\begin{figure}[t]
    \centering
    \vspace{-3mm}
    \begin{overpic}[width=0.63\linewidth, angle=90]{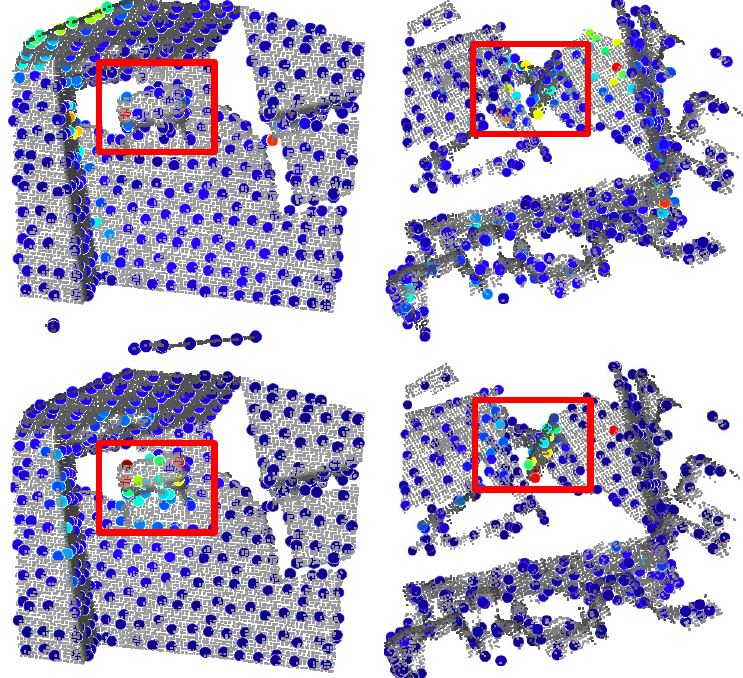}
    \put(15,100){\color{black}\footnotesize\textbf{GeoTrans}}
    \put(65,100)
    {\color{black}\footnotesize\textbf{\ourmethod}}
    \end{overpic}
    \caption{Visualization of attention weights for the point indicated with a red dot. Brighter colors indicate higher attention.}
    \label{fig:attn}
    \vspace{-3mm}
\end{figure}

We conducted an ablation analysis on 3DMatch and 3DLoMatch with \#Samples=1000 to examine the specific roles of each element in our approach. 
To assess the efficacy of geometry information in cross-attention, we compared the registration outcomes of four types of cross-attention - vanilla, pair-wise distance embedding-based (PDE), triplet-wise angle embedding-based (TAE), and the whole geometric cross-attention (DAE) in Table~\ref{tb:abcross}. 
Both pair-wise distance and triplet-wise angle embedding can improve the registration performance. 
To be more detailed, Geometric information improves the performance 
by nearly 0.2\% (92.0\% vs. 92.2\%) RR, 0.1\% (98.0\% vs. 98.1\%) FMR, and 2.8\% (81.4\% vs. 84.2\%) IR on 3DMatch indicating that \ourmethod benefits from pair-wise distance and triplet-wise angle embedding. 
Additionally, the study presents results for employing self-attention mechanisms combined with distance mapping, a technique that further enhanced performance.
Fig.~\ref{fig:attn} presents two examples of attention maps. 
It includes only two source point clouds and omits target point clouds for clarity. 
Given a point from the source, the red boxes highlight the correct correspondence regions. Our network focuses on corresponding superpoints in the source point cloud for each point in the target cloud.
We emphasize that attention is selectively focused on areas within the red rectangles, ensuring clarity on this key aspect. Additionally, the figure contrasts attention maps from two example configurations, showcasing our method’s superior ability to identify overlapping regions compared to GeoTr. 
In these examples, a selected point in the target point cloud is matched with corresponding superpoints in the source point cloud. 
Blue points indicate low attention weights, while points in other colors signify higher attention weights. 
Notably, our approach focuses attention on the red rectangles. 

\section{Conclusions}
We present a fully-geometric attention algorithm to achieve accurate point cloud registration through coarse-to-fine matching.
We fuse information from coordinates and features at the super-point level between point clouds, an unaddressed problem in the literature primarily because it must guarantee rotation and translation invariance as point clouds reside in different and independent reference frames.
Cross-attention can identify overlap areas and accurately match coarse features.
Instead, self-attention produces more distinctive local features for fine matching.
The results showed that our approach achieves state-of-the-art performance on a large benchmark, including 3DMatch, 3DLoMatch, KITTI, and 3DCSR datasets.

\noindent \textbf{Limitations.}
The time complexity of computing the pairwise distance mapping is relatively high. However, this operation acts on superpoints, making it more manageable. 
Addressing the issue of dealing with source and target point clouds of different densities, especially since different sensors capture them, remains a challenge.

\footnotesize\noindent\textbf{Acknowledgment.}~This work was sponsored by the FAIR - Future AI Research (PE00000013), funded by NextGeneration EU.
Bruno Lepri and Nicu Sebe acknowledge funding by the European Union’s Horizon Europe research and innovation program under grant agreement No. 101120237 (ELIAS).
{
    \small
    \bibliographystyle{ieeenat_fullname}
    \bibliography{main}
}

\end{document}